\pdfoutput=1
\documentclass[11pt]{article}

\usepackage[final]{acl}
\usepackage{fvextra}
\usepackage{times}
\usepackage{latexsym}
\usepackage{booktabs}
\usepackage[T1]{fontenc}

\usepackage[utf8]{inputenc}

\usepackage{microtype}

\usepackage{inconsolata}

\usepackage{graphicx}
\usepackage[most]{tcolorbox}

\definecolor{promptgray}{gray}{0.96}

\newtcblisting{promptbox}{
  breakable,
  listing only,
  colback=promptgray,
  colframe=promptgray,
  boxrule=0pt,
  arc=2pt,
  left=5pt,
  right=5pt,
  top=5pt,
  bottom=5pt,
  before skip=6pt,
  after skip=6pt,
  listing options={
    basicstyle=\ttfamily\scriptsize,
    breaklines=true,
    columns=fullflexible,
    keepspaces=true
  }
}

\title{GeopoliticalQA: Benchmarking Retrieval and Generation\\ over Intelligence Reports}

\author{
  \textbf{Veronika Solopova},
  \textbf{Ivan Havlytskyi},
  \textbf{Viktoria Skorik},
  \textbf{Maksym Tereshchenko},
  \textbf{Ostap Vykhopen}
\\
  \small{
    \textbf{Correspondence:}
    \href{mailto:ostap.vykhopen@mantisanalytics.com}
    {ostap.vykhopen@mantisanalytics.com}
  }
}

\begin{document}
\maketitle
\begin{abstract}
Geopolitical and strategic intelligence reports constitute a rich source of expert knowledge and have increasing value in the current political climate, but their heterogeneity and analytical complexity pose challenges for retrieval-augmented question answering. We introduce a benchmark for grounded question answering over 415 open-source geopolitical and strategic intelligence reports, comprising 5,525 document-grounded question–answer pairs spanning factual, inferential, comparative, and risk-oriented questions. We propose a document-aware generation pipeline that profiles reports before generating questions and validate dataset quality through expert annotation and an LLM-as-judge protocol. Using the benchmark, we evaluate 15 retrieval systems and six open-weight LLMs for RAG. We find that models perform similarly on document-level retrieval, while at the harder chunk level retrieval quality is not driven by embedding-model scale: bge-m3, a mid-size open encoder matches an order of magnitude larger ones and outperforms commercial embedding APIs. In contrast, answer generation remains challenging even given the gold evidence, particularly for inferential and risk-oriented questions.
\end{abstract}

\section{Introduction}

In recent years, governments, international organisations, think tanks, humanitarian agencies, and commercial intelligence providers have produced an unprecedented volume of geopolitical and strategic analysis. These reports synthesize information on armed conflicts, foreign information manipulation, sanctions, cyber threats, humanitarian crises, election integrity, supply-chain risks, and other developments relevant to political and security decision making. Although intended primarily for human analysts, collectively they represent a rich and continually growing repository of expert knowledge.  

Large language models (LLMs) and retrieval-augmented generation (RAG) systems offer the possibility of making this knowledge substantially more accessible. Instead of manually searching hundreds of lengthy reports, analysts could query them directly, rapidly retrieve supporting evidence, compare findings across organisations, or integrate information into downstream analytical workflows. Such capabilities could support strategic communications, intelligence analysis, crisis response, policy development, and corporate risk assessment.

However, despite the practical importance of this application, it remains unclear how reliably current LLMs can answer questions grounded in geopolitical intelligence reports. Unlike many existing document QA benchmarks, these reports are highly heterogeneous: they include policy briefs, intelligence assessments, narrative trackers, methodological codebooks, humanitarian reports, and technical analyses, each requiring different forms of reasoning and different notions of factual grounding. Moreover, many reports discuss uncertainty, competing narratives, or future risks rather than objective facts, making conventional question-generation strategies poorly suited to the domain. To the best of our knowledge, no existing benchmark systematically evaluates question answering over this class of documents.

To address this gap, we introduce GeopoliticalQA, a benchmark for grounded question answering over geopolitical and strategic intelligence reports. Rather than relying on manually authored questions, we develop an automated pipeline that first profiles each document to determine what types of questions it can legitimately answer before generating document-appropriate question-answer pairs grounded in retrieved evidence. This allows us to construct a benchmark spanning over 400 reports from diverse organisations while preserving the analytical characteristics of different report types.

Beyond constructing the benchmark, we investigate two practical questions that are central to scalable benchmark creation. First, can LLMs serve as reliable judges for assessing the quality of automatically generated question-answer pairs, reducing the need for expensive expert annotation? Second, can relatively small open-weight language models generate high-quality questions and answers for this domain, enabling reproducible and cost-effective dataset construction?

Our contributions are:\\
1. We introduce the first benchmark for retrieval-based question answering over geopolitical and strategic intelligence reports, comprising 415 reports and 5,525 grounded question-answer pairs spanning diverse document types;\\
2. We propose a document-aware generation pipeline that profiles reports before generating document-appropriate, grounded questions, avoiding common failure modes of fixed-template generation; \\
3. We evaluate the use of a multi-model LLM-as-judge framework for automatic quality assessment and compare it with expert human evaluation; \\
4. We demonstrate that a relatively small open-weight model can generate a high-quality benchmark for this challenging domain, providing a practical methodology for constructing future domain-specific RAG datasets.  

\section{Related Work}

Question answering has expanded from short, fact-oriented passages to long and specialized documents. Benchmarks such as NarrativeQA~\cite{kovcisky2018narrativeqa} and subsequent work on narrative question answering~\cite{mou2021narrative} show that retrieving and integrating evidence across lengthy documents remains challenging. Long-context benchmarks further evaluate these capabilities across diverse tasks and document lengths~\cite{bai2023longbench,zhang2024infinitebench}.

Domain-specific benchmarks increasingly assess retrieval-augmented generation (RAG) in realistic settings. FinanceBench~\cite{islam2023financebench} evaluates question answering over financial reports, \citet{parikh2024esg} develop a retrieval-based QA system for long ESG reports, and \citet{filice-etal-2025-generating} generate Q\&A benchmarks for RAG in enterprise settings. These studies demonstrate the value of domain-grounded evaluation, but only address financial, corporate, or otherwise specialized document collections rather than geopolitical intelligence reports. Geopolitical information presents additional challenges because answers may depend on the selected sources, languages, and represented perspectives. \citet{li-etal-2025-multilingual} introduce a multilingual RAG benchmark covering territorial disputes across 49 languages and show that the linguistic composition of retrieved documents affects response consistency and geopolitical bias, focusing on fact-seeking questions over retrieved Wikipedia passages rather than long-form expert analysis. \citet{dellorto2026genome} use LLMs to extract, classify, normalize, and deduplicate geopolitical events from newswire articles, producing a structured dataset of conflictual and cooperative interactions. Our setting instead comprises heterogeneous intelligence reports that combine factual evidence with interpretation, uncertainty, forecasts, and recommendations, and therefore carry a high risk of hallucination, inconsistency, limited transparency, and geopolitical bias~\cite{thapa-etal-2025-challenges}. Answering questions over such documents thus requires not only retrieving relevant passages but also synthesizing analytical claims distributed across report sections.

\section{Methods}
\subsection{Corpus Collection}
\label{sec:corpus}
We assemble a corpus of geopolitical and intelligence-analysis reports comprising \textbf{415} open-source
institutional publications from 122 distinct publishers, spanning the
institutional categories shown in Table~\ref{tab:corpus}; it is strongly contemporary,
with a median publication year of 2025 and 94\% of documents dated 2022 or later
(range 2003--2026).

\begin{table}[t]
\centering
\small
\begin{tabular}{lr}
\toprule
Institutional category & Documents \\
\midrule
Regional / local think tanks & 129 \\
Specialised domains          & \phantom{0}64 \\
Official \& multilateral     & \phantom{0}63 \\
Global think tanks           & \phantom{0}51 \\
Index \& data-narrative      & \phantom{0}31 \\
Parliamentary services       & \phantom{0}29 \\
Humanitarian                 & \phantom{0}20 \\
Academic policy              & \phantom{0}15 \\
Human rights                 & \phantom{0}13 \\
\midrule
Total            & 415 \\
\bottomrule
\end{tabular}
\caption{Corpus composition by institutional category (122 publishers).}
\label{tab:corpus}
\end{table}

\paragraph{Agent-driven collection.}
The corpus was gathered using LLM agents operating a real browser, one publisher at a time, navigating each institution's own publications pages as a human researcher would (going from landing page to index and to document page 
download). Agents ran on Claude~Opus within a multi-agent harness
(subagent fan-out) using a version-pinned \texttt{browser-use} automation skill
(pinned by content hash for post-hoc reproducibility). Collection occurred in a
single session on 4~May~2026 and is a point-in-time snapshot: there was
no incremental top-up.

\paragraph{Source selection.}
Source selection is specified by a single structured ``research librarian''
prompt (reproduced verbatim in Appendix~\ref{app:prompt}), which is the
load-bearing methodological artifact: it inventories \emph{publisher series},
not individual documents, across eight institutional categories-official and
multilateral bodies, parliamentary research services, global think tanks,
regional and local think tanks (designated the highest-priority category for
diversity), humanitarian and human-rights organisations, index and
data-narrative providers, specialised domains, and academic policy outlets-and
instructs the model to maximise variance along five dimensions: institutional
type, geographic origin, epistemic stance, document genre, and temporal coverage.
This stage produced an inventory of \textbf{126 publishers}, each with a
pre-assigned corpus path; 122 of these ultimately contributed at least one
document to the final corpus (four yielded no retrievable English documents).
Humanitarian and human-rights sources form a single source-selection task but are reported as two categories in Table~\ref{tab:corpus}. Three selection policies are encoded in the prompt and applied \emph{before} retrieval:
(i) language filtering is preventive-agents were pointed only at
English-language publication streams rather than collecting broadly and filtering after the fact; (ii) access screening is preventive-only openly
accessible or registration-only sources were inventoried and hard paywalls were
excluded, so no paywalled material was ever retrieved; and
(iii) bias is recorded, not filtered-each publisher carries explicit
\emph{variance} and \emph{bias/limitations} fields, and advocacy-leaning sources are retained on the principle that bias diversity is itself a corpus value.

\paragraph{Partitioned retrieval and de-duplication.}
Retrieval was fanned out across subagents, each assigned \emph{exactly one
publisher and one disjoint output directory} taken from the inventory. Hence, agents cannot collide because their write targets are disjoint by construction, which also makes the corpus a
publisher taxonomy rather than a topic taxonomy. Document-level
de-duplication is therefore an emergent property of the partition, with known limitations, such as co-publication (a document
legitimately filed under two publishers) and near-duplicates (reissues, summary
vs.\ full report, annual editions with heavy reuse), which were not tested.
(This is distinct from the chunk-level de-duplication applied at indexing time,
\S\ref{sec:indexing}.)

\paragraph{Language and access basis.}
Language control is two-stage: preventive at source selection, and detective
post~hoc (per-document Latin word counts, English stopword ratios, and non-Latin character counts). Residual non-Latin text is almost entirely foreign-language \emph{citations} in otherwise-English documents; a single genuinely mixed English-Russian document warrants language-aware chunking. Every document was retrieved from its publisher's openly accessible page with no authentication, paywall circumvention, or third-party aggregator. 

\paragraph{Human review.}
Collection was verified by a per-publisher spot check: for every publisher
directory at least one document was opened and checked for correct attribution,
English language, and substance (a genuine report, not a landing page or cover
sheet), with every publisher verified at least once (roughly one document in
three).

Document profiling (\S\ref{sec:generation}) further classifies the corpus by
type: the majority are \emph{analytical assessments} (317), followed by
\emph{narrative/disinformation trackers} (45), \emph{annual reports} (18),
\emph{policy briefs} (10), and \emph{methodology codebooks} (4), among others.
This diversity is deliberate: it forces the benchmark to cover documents with
very different affordances.

\subsection{Parsing, Chunking, and Indexing}
\label{sec:indexing}
Source documents (PDF, DOCX, etc.) are converted to Markdown via an OCR/layout pipeline (Mistral OCR\footnote{\url{https://docs.mistral.ai/studio/document-processing/basic_ocr}} with a Docling converter \cite{livathinos2025doclingefficientopensourcetoolkit}) that preserves section structure, then cleaned and normalised before segmentation. 

Chunking is structure-aware and hierarchical, following the document's section tree (chunk depth levels 2--7), and each chunk retains the title of its section, hierarchy path, page number, character span and predecessor/successor links, with a target minimum chunk size of 300 characters and dedicated handling for tables ($\approx$16\% of chunks). The corpus yields 55{,}353 chunks (mean 1{,}138 / median 723 characters, $\approx$285/181 tokens). Chunks carry a content hash; $\approx$2.8\% are exact-duplicate rows (shared boilerplate recurring across reports) and no near-duplicate removal is applied. Chunks are embedded with \texttt{intfloat/multilingual-e5-large-instruct} and stored in a Milvus vector database.

Retrieval is hybrid: a dense branch (cosine similarity over the e5
embeddings) and a sparse branch (BM25) are fused with Reciprocal Rank Fusion
(RRF, $k{=}60$), returning the top $k{=}8$ chunks. The same retriever is used
both to construct the benchmark and as the system-under-test, allowing end-to-end evaluation of retrieval quality.

\subsection{Q\&A Generation}
\label{sec:generation}
Our generation pipeline is document-grounded and content-gated, designed
to avoid a failure mode observed in early iterations: a fixed
persona\,$\times$\,question grid forces analytically inappropriate questions
(e.g., asking a methodology codebook about oil prices). The pipeline runs two LLM stages per document (all stage prompts: profiling, question generation, and answering, are reproduced in Appendix~\ref{app:genprompts}). These step use \textbf{Gemma}\footnote{\texttt{https://huggingface.co/google/gemma-4-26B-A4B-it}} served via vLLM (temperature $0.2$). We deliberately self-host on 1$\times$H100 rather than call a hosted
API: geopolitical and humanitarian content triggers frequent false-positive
content-moderation refusals on commercial endpoints, which would silently bias
the corpus toward ``safe'' documents.

\paragraph{(a) Profiling.}
For each document we issue eight semantic \emph{probe} queries (scope, key
findings, actors, locations, timeline, risks, recommendations, quantitative
claims) and retrieve the top-4 chunks each, yielding a representative chunk set.
A single LLM call then produces a structured profile: the document \emph{type}
and \emph{purpose}, and explicit \texttt{can\_answer}/\texttt{cannot\_answer}
scopes.

\paragraph{(b) Generation.}
A single chunk-grounded call per document receives the representative chunks, the
profile, and a catalogue of 16 analyst personas (e.g., \emph{Geopolitical
Researcher}, \emph{Intelligence Analyst}, \emph{Conflict Data Analyst},
\emph{Cyber Threat Analyst}, \emph{Sanctions Compliance Analyst}). The model
(i)~selects only the personas the \emph{content} supports, and (ii)~emits up to 20
atomic questions, each labelled with one of four types (\textbf{factual},
\textbf{inferential}, \textbf{comparative}, \textbf{risk-oriented}) together with
a rationale and a grounding pointer to the source chunks. The prompt enforces
rules distilled from expert review, such as one atomic ask per question, no forecast framing on retrospective documents, comparisons only between entities present in-text, and claims from narrative sources phrased as \emph{reported} rather than \emph{asserted}. A post-processing step then removes near-duplicate questions (Jaccard overlap of content tokens).

\paragraph{(c) Answering.}
Each question is re-retrieved (hybrid, top-8) and answered by an LLM constrained
strictly to the retrieved chunks, with inline chunk citations, modal hedging on
inferential/risk items, and an explicit \texttt{INSUFFICIENT\_CONTEXT} response
when the evidence is inadequate. Both the answer and its ground-truth retrieved
chunks are stored, so each item can evaluate retrieval and answer generation.

The resulting benchmark contains \textbf{5{,}525} Q\&A pairs over 413 documents
(2 of the 415 collected reports yielded no retrievable chunks and were skipped),
averaging $13.4$ questions per document. Question types follow the natural
distribution in Table~\ref{tab:qtypes}.

\begin{table}[t]
\centering
\small
\begin{tabular}{lrr}
\toprule
Question type & Count & \% \\
\midrule
Factual        & 3{,}805 & 68.9 \\
Inferential    & 789     & 14.3 \\
Comparative    & 696     & 12.6 \\
Risk-oriented  & 235     & \phantom{0}4.3 \\
\midrule
Total          & 5{,}525 & 100.0 \\
\bottomrule
\end{tabular}
\caption{Question-type distribution of the benchmark.}
\label{tab:qtypes}
\end{table}

\subsection{Validation}
\label{sec:validation}
We assess quality on six dimensions, each on a $1$--$3$ scale
($3$~excellent, $2$~acceptable but flawed, $1$~poor/broken):
\emph{groundedness} (is the question answerable from the chunks),
\emph{document-purpose fit}, \emph{atomicity}, \emph{clarity},
\emph{answer faithfulness}, and an \emph{overall} judgment. We first collect human annotations on a sample, then select and validate an LLM judge against them, and finally apply that judge to the full corpus (its scores are reported in \S\ref{sec:res-quality}).

\paragraph{Human annotation.}
Three domain analyst annotators each rate
$\sim$\textbf{150 items}: a shared 100-item overlap block (identical
across annotators, for inter-annotator agreement) and $\sim$50 unique items,
giving 250 distinct annotated items, mapped to a dataset row
and used for judge validation and the RAG evaluation). Items are drawn from
the corpus, stratified to preserve the natural question-type distribution, and
scored strictly against the retrieved chunks following a written rubric with
per-dimension anchors. Inter-annotator agreement on the overlap block is reported in Table~\ref{tab:iaa}. Raw agreement is high (majority agreement 99\%; exact three-way agreement 63\%), but chance-corrected agreement is modest (mean quadratic-weighted Cohen's $\kappa{=}0.24$): the score distribution is top-heavy (most items are rated $3$), which deflates $\kappa$, and one annotator is systematically stricter. Annotators thus agree strongly on the aggregate quality and on \emph{which} items are sound, while disagreeing mainly at the \emph{good} vs.\ \emph{acceptable} boundary.

\begin{table}[t]
\centering\small
\begin{tabular}{lccc}
\toprule
Dimension & Cohen's $\kappa$ & exact (3/3) & $\geq$2/3 \\
\midrule
Groundedness    & 0.42 & 78\% & 99\% \\
Faithfulness    & 0.30 & 58\% & 100\% \\
Overall         & 0.28 & 54\% & 98\% \\
Clarity         & 0.28 & 60\% & 98\% \\
Doc-purpose fit & 0.12 & 71\% & 99\% \\
Atomicity       & 0.05 & 60\% & 99\% \\
\midrule
Mean            & 0.24 & 63\% & 99\% \\
\bottomrule
\end{tabular}
\caption{Inter-annotator agreement on the 100-item overlap block (three
annotators). $\kappa$ is quadratic-weighted, averaged over rater pairs.}
\label{tab:iaa}
\end{table}

\paragraph{Judge selection and validation.}
We score the 250 annotated items with three candidate judges from different
families: \textbf{GPT-5.4}, \textbf{Claude~Sonnet~5}~\citep{claude_sonnet_5},
and \textbf{DeepSeek-V3.2}~\citep{deepseekai2025deepseekv32pushingfrontieropen} (none of which is the
generator) each applying the identical $1$--$3$ rubric, and compare them to the
human labels (Table~\ref{tab:judgesel}). \textbf{GPT-5.4} agrees best with the human
annotators: pooled across dimensions it matches the two consensus annotators at
$80\%$ and $75\%$ exact agreement (and the strict outlier at $63\%$), with a
Spearman correlation up to $0.43$ on \emph{overall}, matching the strongest
human--human pair ($0.42$). Its agreement with humans thus falls \emph{within the
human inter-annotator range}: the judge agrees with the consensus annotators about as well as they agree with each other, and better than the outlier annotator agrees with anyone. We therefore adopt GPT-5.4 as the automatic judge. The full judge prompt is reproduced in Appendix~\ref{app:judge})

\begin{table}[t]
\centering\small
\begin{tabular}{lccc}
\toprule
Agreement with $\rightarrow$ & A1 & A2 & A3 \\
\midrule
GPT-5.4         & \textbf{80\%} & 63\% & \textbf{75\%} \\
Claude Sonnet 5 & 75\% & 59\% & 71\% \\
DeepSeek-V3.2   & 76\% & 61\% & 71\% \\
\midrule
\textit{other annotators} & 78\% & 69\% & 79\% \\
\bottomrule
\end{tabular}
\caption{Judge validation: pooled exact agreement (all six dimensions) \emph{with}
each human annotator (A1--A3). Judge rows give judge-vs-annotator agreement; the
\emph{other annotators} row gives the mean agreement of the remaining two
annotators with that annotator (the per-annotator human ceiling). A2 is the
systematically stricter annotator, so all raters (human and model) agree with A2
least.}
\label{tab:judgesel}
\end{table}

\paragraph{Scaling the judge to the full split.}
To score all 5{,}525 items economically, we run GPT-5.4 with
\texttt{reasoning\_effort=low} via the OpenAI Batch API. We verified on the 250
annotated items that low-effort scoring is statistically equivalent to the default
(identical exact-agreement and macro-$F_1$ with humans; overall mean unchanged at
$2.5$), so the cost reduction does not compromise the metric. Judging ran on the
standard OpenAI endpoint (not the moderated deployment that affected generation) and
returned valid scores for all 5{,}525 items; no judge call was refused or dropped.
The resulting dataset-quality scores are reported in Table~\ref{tab:quality}.

\section{Results}
\label{sec:results}

We report on 
5{,}525 Q\&A pairs, of which the cleaned primary benchmark comprises
\textbf{4{,}980} items (every dimension rated $\geq 2$ by the validated judge) and a stricter \emph{gold} tier comprises 3{,}503 items (all dimensions rated 3).

\subsection{Dataset Quality}
\label{sec:res-quality}
Table~\ref{tab:quality} reports the automatic quality of the generated data,
scored by GPT-5.4 on the 1--3 rubric that our human study validated
(\S\ref{sec:validation}). Quality is high overall (mean 2.60/3; 66\% of items
rated \emph{excellent}), and the guardrails introduced at generation time hold at scale with high averages for \emph{document-purpose fit} (2.99), \emph{clarity} (2.97) and
\emph{atomicity} (2.92). \emph{Faithfulness} (2.60) is the
weakest dimension and the only one with a non-trivial broken rate (8\%),
consistent with the human annotators' findings.

\begin{table}[t]
\centering\small
\begin{tabular}{lccc}
\toprule
Dimension & Mean & \% rated 3 & \% broken (1) \\
\midrule
Doc-purpose fit & 2.99 & 99 & 0 \\
Clarity         & 2.97 & 97 & 0 \\
Atomicity       & 2.92 & 93 & 0 \\
Groundedness    & 2.75 & 82 & 6 \\
Faithfulness    & 2.60 & 69 & 8 \\
\midrule
Overall         & 2.60 & 66 & 6 \\
\bottomrule
\end{tabular}
\caption{Dataset quality (GPT-5.4 judge, 1--3;
$n{=}5{,}525$).}
\label{tab:quality}
\end{table}

\subsection{Retrieval Benchmark}
\label{sec:res-retrieval}
We evaluate 15 retrieval systems: a lexical baseline (BM25), open dense encoders across four families and multiple sizes, and three commercial embedding APIs. They are tested on the 4{,}980 clean-tier queries against the 55{,}353-chunk corpus. We report \emph{document-level} retrieval (does a chunk of the source document appear in the top-$k$; lenient) and \emph{chunk-level} retrieval against the \emph{answer-cited} chunks (the passages the reference answer actually cites (mean $1.55$ per item), which we take as the answer-bearing gold); chunks with an identical content hash to a gold chunk are credited. Table~\ref{tab:retrieval} gives the full leaderboard.

\begin{figure}[t]
\centering
\includegraphics[width=\columnwidth]{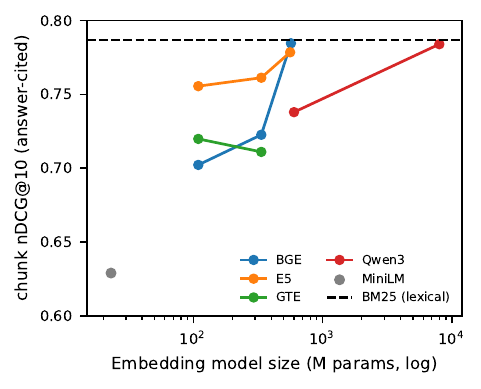}
\caption{In-domain retrieval scaling: chunk nDCG@10 (answer-cited gold) vs.\
encoder size (log scale) by family. Dense encoders largely scale with size (GTE is
an exception, with gte-base above gte-large), and the best (bge-m3, Qwen3-Emb-8B)
match the BM25 baseline (dashed).}
\label{fig:scaling}
\end{figure}

Document-level retrieval performance shows similar trend across models: every system reaches $\geq 0.93$ R@10, so the task does not
discriminate models at the document granularity. On chunk-level nDCG@10
among independently evaluated systems, bge-m3 (0.785) and Qwen3-Embedding-8B (0.784) lead, and their difference is not significant (paired bootstrap $\Delta{=}0.001$, 95\% CI $[-0.007, 0.008]$); BM25 (0.787), though in-construction, is likewise statistically indistinguishable from bge-m3 ($\Delta{=}0.002$, CI $[-0.007, 0.011]$). Notably, bge-m3 (568M) matches Qwen3-Embedding-8B despite being an order of magnitude smaller, while significantly outperforming every commercial API (e.g.\ vs.\ OpenAI-3-large, $\Delta{=}0.068$, CI $[0.059, 0.076]$; all paired CIs from 5{,}000 bootstrap resamples over the 4{,}856 cited-gold queries). Finally, dense models show in-domain size scaling within most families (Figure~\ref{fig:scaling}), but not universally: gte-base slightly yet significantly exceeds the larger gte-large ($\Delta{=}0.009$, CI $[0.002, 0.016]$). The commercial APIs (OpenAI, Mistral) land only mid-pack.

\begin{table*}[t]
\centering\small
\begin{tabular}{llccccc}
\toprule
System & Size & doc R@1 & doc R@10 & chunk R@10 & chunk nDCG@10 & chunk MRR@10 \\
\midrule
BM25$^\ast$ & --- & 0.840 & 0.974 & 0.877 & 0.787 & 0.801 \\
bge-m3        & 568M & 0.836 & 0.974 & 0.885 & \textbf{0.785} & 0.793 \\
Qwen3-Emb-8B  & 8B   & 0.839 & 0.975 & 0.892 & 0.784 & 0.788 \\
me5-large$^\ast$   & 560M & 0.831 & 0.974 & 0.884 & 0.778 & 0.786 \\
e5-large-v2$^\ast$ & 335M & 0.828 & 0.970 & 0.870 & 0.761 & 0.768 \\
e5-base-v2$^\ast$  & 109M & 0.811 & 0.969 & 0.868 & 0.756 & 0.761 \\
Qwen3-Emb-0.6B& 0.6B & 0.815 & 0.963 & 0.854 & 0.738 & 0.743 \\
bge-large     & 335M & 0.808 & 0.963 & 0.831 & 0.723 & 0.735 \\
gte-base      & 109M & 0.792 & 0.957 & 0.836 & 0.720 & 0.726 \\
OpenAI-3-large& API  & 0.777 & 0.944 & 0.845 & 0.717 & 0.715 \\
gte-large     & 335M & 0.792 & 0.956 & 0.825 & 0.711 & 0.718 \\
bge-base      & 109M & 0.788 & 0.955 & 0.816 & 0.702 & 0.713 \\
mistral-embed & API  & 0.778 & 0.944 & 0.831 & 0.701 & 0.699 \\
OpenAI-3-small& API  & 0.769 & 0.941 & 0.834 & 0.700 & 0.695 \\
MiniLM-L6     & 23M  & 0.738 & 0.930 & 0.743 & 0.629 & 0.637 \\
\bottomrule
\end{tabular}
\caption{Retrieval leaderboard ($n{=}4{,}980$ queries, 55{,}353 chunks), ranked by
chunk nDCG@10 against answer-cited gold. Query/passage prefixes follow each
family's recommended format (Appendix~\ref{app:embprefix}). $^\ast$\,in-construction (BM25 and \texttt{e5} were
the construction hybrid's components); shown for reference, not ranked as winners.}
\label{tab:retrieval}
\end{table*}

\subsection{RAG Reader Benchmark}
\label{sec:res-rag}
We evaluate six open readers (Llama, Qwen, Mistral, at small and mid sizes) as the generation stage on the 250 annotated items (a stratified external sample, not
restricted to the clean tier). To isolate generation from retrieval, each reader is
given the gold retrieved context and answers the question; the judge then scores
each answer given the question, the gold context, and the reference answer, on
\emph{faithfulness}, \emph{correctness}, and \emph{overall}
(Table~\ref{tab:readers}). \emph{Faithfulness} and \emph{overall} are among the
dimensions validated against humans (\S\ref{sec:validation}); \emph{correctness} is
an additional dimension we report as exploratory, and because the reference answers
are themselves imperfect ($\approx$8\% judged unfaithful, \S\ref{sec:res-quality})
it should be read as agreement-with-reference rather than ground truth.

Mid-tier models outperform small ones (mean overall 2.53 vs.\ 2.35). The top two,
Llama-3.3-70B (2.56) and Mistral-Small-3.2 (2.55), do not have statistical
difference (overlapping 95\% bootstrap CIs; paired-difference CI
$[-0.11, 0.13]$ spans zero), so we do not rank them. Qwen3-8B attains the highest
\emph{correctness} (2.66) despite its size. Small models are competitive on
answer content, but they refuse far more often
(\S\ref{sec:error}). Bootstrap 95\% CIs (5{,}000 resamples) are shown for overall.

\begin{table}[t]
\centering\scriptsize
\begin{tabular}{llccc}
\toprule
Reader & Size & Faith. & Correct. & Overall [95\% CI] \\
\midrule
Llama-3.3-70B      & 70B & 2.69 & 2.60 & 2.56 [2.48, 2.64] \\
Mistral-Small-3.2  & 24B & 2.60 & 2.60 & 2.55 [2.46, 2.63] \\
Qwen3-8B           & 8B  & 2.46 & 2.66 & 2.48 [2.40, 2.56] \\
Qwen3-32B          & 32B & 2.48 & 2.67 & 2.48 [2.40, 2.56] \\
Llama-3.1-8B       & 8B  & 2.48 & 2.34 & 2.30 [2.20, 2.39] \\
Mistral-Nemo       & 12B & 2.57 & 2.30 & 2.28 [2.17, 2.38] \\
\midrule
\textit{small mean}& --- & & & 2.35 \\
\textit{mid mean}  & --- & & & 2.53 \\
\bottomrule
\end{tabular}
\caption{RAG reader leaderboard (GPT-5.4 judge, 1--3; $n{=}250$), readers given
gold context to isolate generation quality. Faithfulness and overall are validated
dimensions; correctness is exploratory.}
\label{tab:readers}
\end{table}

\section{Error Analysis}
\label{sec:error}
We analyse the best system on each task: BM25 (retrieval) and Llama-3.3-70B
(generation).

Even BM25 places at least one answer-cited chunk in the top-10 for
93--96\% of queries and the source document for $\sim$97\%,
near-uniformly across question types (Table~\ref{tab:reterr}). The spread in
chunk nDCG@10 across systems therefore reflects \emph{completeness}, recovering
\emph{all} of the (on average 1.55) answer-cited chunks of an item, rather than a failure to find anything relevant: per-query recall of the cited set ranges from
89\% (factual) down to 78\% (risk-oriented). Residual misses include e.g. fine-grained rubric definitions, fieldwork specifics and are not type-driven.

\begin{table}[t]
\centering\small
\begin{tabular}{lcccc}
\toprule
Question type & $n$ & doc-miss & chunk-miss & recall \\
\midrule
Factual       & 3416 & 3\% & 7\% & 89\% \\
Inferential   & 719  & 3\% & 6\% & 82\% \\
Comparative   & 626  & 2\% & 6\% & 88\% \\
Risk-oriented & 219  & 3\% & 4\% & 78\% \\
\bottomrule
\end{tabular}
\caption{BM25 retrieval failures at top-10 by question type, against
\emph{answer-cited} gold (with content-hash dedup credit): fraction of queries
retrieving no source-document chunk (doc-miss) / no answer-cited chunk
(chunk-miss), and mean per-query recall of the cited set.}
\label{tab:reterr}
\end{table}

\begin{figure}[t]
\centering
\includegraphics[width=\columnwidth]{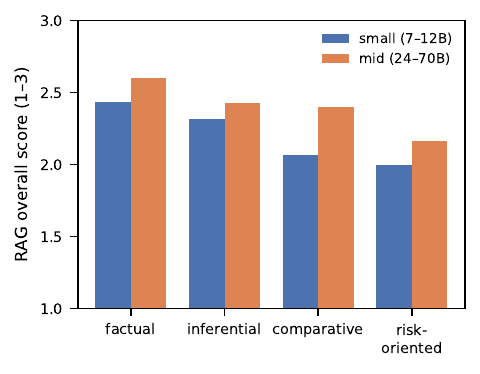}
\caption{RAG answer quality (overall, 1--3) by question type and model size.
Both tiers degrade from factual to inferential to risk-oriented questions; the
small--mid gap is largest on the harder, reasoning-heavy types.}
\label{fig:readererr}
\end{figure}

Given gold context, the best reader still fails on a substantial share of items,
and the failure rate scales with reasoning demand: 28\% of \emph{factual} items are
flawed (overall $\leq 2$), rising to 38\% \emph{comparative}, 47\% \emph{inferential}
and 70\% \emph{risk-oriented} (Figure~\ref{fig:readererr}; the figure orders types by
the taxonomy, so \emph{comparative} appears after \emph{inferential} despite its lower
flawed rate). Among the \emph{broken}
(score-1) items, three failure modes recur: \emph{hallucination} (unsupported
facts/figures; 9\%, faithfulness $=1$), \emph{incompleteness/incorrectness} (7\%,
correctness $=1$ on non-refusals), and \emph{unjustified refusal} (2\%; answering
\texttt{INSUFFICIENT\_CONTEXT} when the evidence is present). These modes count
only score-1 items and therefore do not decompose the larger $\leq 2$ flawed rates,
which also include score-2 minor issues. Refusal is strongly size-dependent, with small
readers refusing more often (Mistral-Nemo 18\%, Llama-3.1-8B 14\%, mostly unjustified),
whereas mid-tier readers refusing $\leq 5\%$ and the Qwen models $\leq 1\%$.

\section{Discussion and Conclusion}
In this paper, we created a benchmark for QA and information retrieval over geopolitical and strategic intelligence reports and evaluated popular retrievers and LLMs for RAG performance. Our findings challenge the assumption that stronger semantic retrieval necessarily requires increasingly large embedding models. Document-level retrieval performance is similar across the evaluated systems, and at the more demanding chunk level retrieval quality is not driven by embedding-model scale. A mid-size open encoder (bge-m3, 568M) matches a dense model an order of magnitude larger and significantly outperforms commercial embedding APIs. A lexical BM25 baseline is also highly competitive at the chunk level; because BM25, together with e5, was a component of the hybrid used to construct the gold chunks, we report it for reference rather than ranking it as a winner, but its strength is consistent with an important characteristic of the domain: geopolitical reports and analyst questions are highly entity-dense, frequently sharing distinctive names of actors, organisations, programmes, locations, and dates. At the same time, the relatively lower chunk-level recall and nDCG indicate that retrieving some relevant evidence is considerably easier than recovering the complete set of passages needed to support an answer.

Our experiments suggest that generation rather than first-stage retrieval remains the main challenge for grounded QA in this setting. Even when LLMs receive gold evidence, answer quality remains low, and performance deteriorates systematically as questions require more reasoning: 28\% of factual answers produced by the strongest LLMs are flawed, compared with 38\% comparative, 47\% inferential and 70\% risk-oriented. Error analysis shows that these errors include unsupported claims, incomplete or incorrect answers, and refusals to answer despite sufficient available evidence. Providing relevant evidence therefore does not eliminate the difficulty of questions that require going beyond direct factual extraction. These findings are particularly consequential for geopolitical analysis, where the distinction between what a source explicitly reports and what can reasonably be inferred from it is central: a system may retrieve the correct evidence yet introduce unsupported conclusions during synthesis, potentially distorting the resulting analysis.

Additionally, our results also support LLM-based quality assessment for scalable benchmark construction. The selected judge performs within the range of human agreement, although modest inter-annotator agreement ($\kappa$ = 0.24) underscores the continued importance and complexity of human validation in such subjective and cognitively demanding tasks.

Overall, across 415 reports and 5,525 generated Q\&A pairs, our experiments reveal a consistent pattern: identification of relevant source material is often achievable with comparatively simple retrieval methods, whereas choosing the right chunk within the document and faithfully transforming that evidence into an answer remains an open challenge, particularly pronounced when questions require inference, comparison, or risk assessment rather than direct factual extraction. These observations point to several directions for future work. On the retrieval side, improvements may come less from replacing lexical retrieval with larger dense encoders than from improving multi-passage completeness, for example through hybrid retrieval, reranking, query decomposition, or evidence aggregation. On the generation side, the substantial remaining gap even under gold-context conditions suggests a need for methods that explicitly control attribution, uncertainty, and the distinction between reported evidence and model inference.

\section*{Limitations}
Several limitations should be considered. The corpus is English-language and strongly contemporary, with 94\% of the documents published since 2022, limiting conclusions about multilingual geopolitical analysis or historical collections. Its composition is also shaped by the source-selection and accessibility criteria used during collection. Hence, the corpus should not be interpreted as a representative sample of geopolitical analysis globally. In addition, question generation and answering are themselves model-mediated, and the distribution of question types—particularly the predominance of factual questions—reflects both document affordances and the generation pipeline. Finally, although automatic quality assessment is validated against domain analysts, the relatively small human-annotated sample and modest inter-annotator agreement leave room for more extensive expert evaluation, particularly for inferential and risk-oriented questions. The annotators are also the paper's own authors and colleagues, so the validation assesses the output of a pipeline they helped design; this may bias quality judgements upward, and independent third-party annotation would provide a stronger check.

Our evaluation is also single-run: readers are decoded once per item and each answer is scored by a single validated judge rather than a multi-judge panel, so we report bootstrap confidence intervals over items but not over decoding or judge stochasticity. Gold is defined by the chunks an LLM-written reference answer cites, which inherits any citation bias of the generator; we mitigate but do not eliminate this through content-hash dedup credit and the reported by-type recall. We evaluate first-stage retrieval and single-pass reading, and do not study rerankers, iterative or agentic retrieval, or long-context readers that ingest many passages.

\section*{Ethics Statement}
All documents in the corpus are publicly available reports published by their originating organisations; we collected them from public sources. We do not republish the documents themselves: we provide per-document publisher attribution and links to the originals, share the collection with researchers on request, and will remove any item at a rights holder's request. The corpus contains analysis of armed conflict, humanitarian crises, and human-rights situations; it reproduces the framing and potential biases of its sources, and it should be treated as source material for research on retrieval and generation, not as verified ground truth about the events described. No personal data about private individuals was collected, and no human subjects were involved beyond the paper's own authors and colleagues, who annotated a sample of items as part of their professional work with informed consent and were not otherwise compensated as study participants. Question generation and answering use commercial and open-weight LLMs; because these models can produce fluent but unsupported claims, we caution against deploying systems trained or evaluated on this benchmark in operational decision-making without human oversight, and our error analysis is intended to make these failure modes visible. Automated content filters on some commercial APIs refused a portion of this geopolitical material; to avoid biasing the dataset toward filter-permissible content we generated with a self-hosted open model.

\section*{Acknowledgment}
This work was funded by the European Union under the Horizon Europe programme through the European Innovation Council (EIC), grant agreement No. 101311351 (MANTIS4EUR).
\bibliography{custom}

\appendix
\section{Data Availability}
We release our derived artifacts in full for research use: the derived chunks, the 5,525 Q\&A pairs with their question-type and persona metadata, the answer-cited gold chunks, and the per-item LLM-judge scores, together with the retrieval and reader outputs needed to reproduce every table and figure. These are released under CC-BY-4.0. We do not republish the source PDFs themselves, because the 122 publishers' terms vary and many reserve rights even for freely downloadable material; instead, each item carries its original publisher attribution and a link to the authoritative version, and we share the full document collection with researchers on request. All items are external, publicly sourced material; no proprietary or internal data is included. Because the underlying documents are public, we note the usual contamination caveat: portions may appear in the pretraining data of current LLMs, so results on this benchmark should be read as measuring retrieval and grounded synthesis rather than closed-book recall.
\section{Stage-1 Source-Selection Prompt}
\label{app:prompt}
The corpus source-selection methodology is fully specified by the following
prompt, reproduced verbatim; cite or paraphrase from it rather than
reconstructing the criteria from the resulting inventory.
The following prompt was used to identify and evaluate candidate publication sources for corpus construction. We reproduce it in full because it constitutes the source-selection methodology.

\begin{promptbox}
You are a research librarian and geopolitical analyst building a
comprehensive, high-variance English-language document corpus for a
Q&A dataset and benchmark covering all geopolitical domains and
countries. Your task is to inventory and evaluate primary publication
sources -- not individual documents, but publisher series and
collections that produce downloadable files (PDF, DOCX, MD, XLSX,
or similar) on a recurring basis.

---
SCOPE AND GOALS

Cover every geopolitical domain including but not limited to: armed
conflict and war, sanctions and economic coercion, state fragility
and governance, elections and democratic backsliding, human rights
and atrocities, displacement and migration, energy and resource
geopolitics, climate-security nexus, nuclear and WMD
non-proliferation, cyber and information warfare, great power
competition, regional integration and bloc dynamics, debt and
development finance, peacebuilding and transitional justice,
maritime and territorial disputes, counterterrorism, and intelligence
and covert action.

Cover all world regions with deliberate attention to underrepresented
areas: Sub-Saharan Africa, Middle East and North Africa, South Asia,
Southeast Asia, Central Asia, the Caucasus, Latin America and the
Caribbean, Pacific Island states, and the Arctic.

English-language documents only, but actively seek sources that are
non-Western in institutional origin -- local think tanks, regional
intergovernmental bodies, national research institutes, university
policy centres, and parliamentary services outside the US and UK.

File formats in scope: PDF reports, working papers, policy briefs,
DOCX publications, markdown documentation, XLSX datasets with
narrative documentation, and any other downloadable structured
document.

Maximize variance across five dimensions:
1. Institutional type -- multilateral bodies, governments, think
   tanks, academia, NGOs, human rights groups, diaspora institutes,
   military/security establishments, development banks, index
   providers
2. Geographic origin -- balance Global North and Global South
   institutional voices
3. Epistemic stance -- official/diplomatic, technocratic/analytical,
   advocacy/normative, field/operational, quantitative/index-based
4. Document genre -- situation reports, annual outlooks, country
   diagnostics, policy papers, working papers, legal analyses, field
   assessments, oral evidence transcripts, budget documents, peace
   agreements with commentary, incident databases with narratives
5. Temporal coverage -- sources with deep archives (pre-2000) as
   well as high-frequency current publishing (monthly or quarterly)

---
RESEARCH TASKS

Task 1 -- Official and multilateral sources
Inventory UN agencies, regional intergovernmental organizations, and
international financial institutions that publish English-language
reports. For each: confirm the publications landing page URL,
document series names, formats available, publishing cadence, and
whether access is open or paywalled. Include bodies that are rarely
collected: IGAD, ECOWAS, CARICOM, SCO, SAARC, Pacific Community,
ASEAN Secretariat, Eurasian Development Bank, Caribbean Development
Bank, West African Development Bank, and similar.

Task 2 -- Parliamentary and governmental research services
Go beyond CRS and the UK Commons Library. Find equivalent services
in Australia (APH), Canada (Library of Parliament), the EU (EPRS),
India (PRS Legislative Research), South Africa, Kenya, Nigeria,
Israel, Brazil, Germany (Bundestag Wissenschaftliche Dienste),
France, Japan (NDL), and any others with open English-language
publication streams. Confirm what they publish in English and which
series are downloadable.

Task 3 -- Global think tanks and policy research institutes
Cover the full spectrum from the most cited (Brookings, RAND,
Chatham House, IISS, SWP, SIPRI, ICG, CFR, Carnegie, Wilson Center,
Atlantic Council) down to specialized and less-collected
institutions: Peace Research Institute Oslo (PRIO), Clingendael,
NUPI, DIIS, CSS ETH Zurich, GCSP, Geneva Academy, BICC, Stimson
Center, US Institute of Peace, EastWest Institute, Kennan Institute,
and others. Note their publication series, peer-review status, and
funding transparency where known.

Task 4 -- Regional and local think tanks (highest priority for
variance)
This is the most important task for dataset diversity. For each major
world region, find 3--5 credible local or regional think tanks that
publish in English. Examples to verify and expand upon:
- Africa: ISS Africa, African Centre for the Constructive Resolution
  of Disputes (ACCORD), Afrobarometer, Rift Valley Institute, Enough
  Project, Africa Policy Research Institute, Wits P&DM, SAIIA
- Middle East and North Africa: Arab Reform Initiative, Tahrir
  Institute, Chatham House MENA, Carnegie Middle East, Al-Shabaka,
  Arab Center Washington DC, CARPO, Sanaa Center
- South Asia: ORF India, IPCS, Stimson South Asia, SDPI Pakistan,
  CPD Bangladesh, IIDS Nepal, ICES Sri Lanka
- Southeast Asia: ISEAS-Yusof Ishak, RSIS Singapore, CSIS Indonesia,
  ISIS Malaysia, IISS Asia, PacNet (Pacific Forum)
- Central Asia and Caucasus: OSCE Academy Bishkek, CABAR.asia,
  Caucasus Research Resource Centers, Tbilisi-based institutes,
  KAZISS, Friedrich Ebert Stiftung Central Asia offices
- Latin America and Caribbean: CEBRI, Igarape Institute, FLACSO,
  FOCAL, Caribbean Policy Research Institute, Inter-American
  Dialogue, WOLA
- Pacific: PNG National Research Institute, Pacific Community, Lowy
  Institute Pacific programs, PIFS
For each: confirm English-language document output, publication
series, website and download page, formats, and cadence.

Task 5 -- Humanitarian, human rights, and migration sources
Inventory operational and advocacy publishers: ACAPS, REACH
Initiative, Mixed Migration Centre, Internal Displacement Monitoring
Centre (IDMC), Geneva Call, Overseas Development Institute (ODI),
Development Initiatives, FEWS NET, Amnesty International, Human
Rights Watch, FIDH, Article 19, Freedom House, Global Witness,
Global Initiative Against Transnational Organized Crime (GI-TOC),
and any others with downloadable report series.

Task 6 -- Index, dataset, and data-narrative sources
Find sources that combine quantitative indices with downloadable
narrative documentation: V-Dem, Varieties of Democracy working
papers, BTI Transformation Index, Freedom House reports,
Transparency International CPI reports, International IDEA democracy
assessments, AidData, ACLED conflict data narratives, Armed Conflict
Location & Event Data working papers, Global Terrorism Index (IEP),
Nuclear Threat Initiative index documentation, Fragile States Index
(FFP), Economist Intelligence Unit Democracy Index methodology
papers, and similar.

Task 7 -- Specialized and niche domains
Find high-quality English-language publishers in domains that are
often underrepresented in geopolitical corpora:
- Nuclear and WMD: Arms Control Association, Bulletin of the Atomic
  Scientists, James Martin CNS, VERTIC, Middlebury Institute
- Cyber and information operations: Atlantic Council DFRLab,
  Stanford Internet Observatory, Oxford Internet Institute, NATO
  CCDCOE, Australian Strategic Policy Institute (ASPI) cyber
- Energy geopolitics: Oxford Institute for Energy Studies, OIES,
  Chatham House Energy, IEEFA, Ember, Energy Policy Research Group
  Cambridge
- Maritime and territorial: AMTI CSIS, IISS, United Nations Division
  for Ocean Affairs, Stimson maritime
- Transitional justice and accountability: ICTJ, Justice Info, IJRC,
  Global Justice Monitor
- Sanctions and economic statecraft: Castellum.AI methodology docs,
  CSIS Economics, Atlantic Council GeoEconomics Center, OFAC
  methodology publications
- Disinformation and influence operations: EU DisinfoLab, DFRLab,
  GLOBSEC, Hybrid CoE Helsinki

Task 8 -- Academic policy journals and open-access series
Find academic outlets that publish long-form downloadable
policy-oriented papers in open access: Journal of Conflict Resolution
data supplements, Small Wars Journal, Journal of Peace Research data
papers, Security Studies, Global Governance, Stability:
International Journal of Security and Development, African Security,
Journal of Human Security, Third World Quarterly selected open
papers, and equivalent outlets.

---
OUTPUT FORMAT

For each source, provide:
| Field | Content |
|---|---|
| Name | Full institutional name and common abbreviation |
| Type | Institutional category |
| Geographic focus | Primary region(s) covered |
| Thematic focus | Primary topic domains |
| Publications landing page | Direct URL |
| Key document series | Named series and their genre |
| Formats | PDF / DOCX / XLSX / MD / other |
| Cadence | Monthly / quarterly / annual / irregular |
| Access model | Open / registration / partial paywall |
| Archive depth | Earliest documents available |
| Epistemic stance | Official / analytical / advocacy /
  operational / quantitative |
| Variance value | Why this source adds distinct coverage not
  duplicated elsewhere |
| Bias / limitations | Known funding dependencies, national framing,
  selectivity |
| Suggested corpus path | e.g., pdfs/regional/africa/iss_africa/ |
| Priority score | 1--5 (5 = highest yield for Q&A diversity) |

Group sources into the following sections: Official & Multilateral ->
Parliamentary & Governmental -> Global Think Tanks -> Regional &
Local Think Tanks (by sub-region) -> Humanitarian & Human Rights ->
Index & Data-Narrative -> Specialized Domains -> Academic Policy.

At the end, provide:
1. A gap analysis -- regions, domains, or epistemic stances still
   underrepresented after the full inventory
2. A collection sequencing recommendation -- which sources to
   prioritize in a first crawl pass vs. second pass, based on yield,
   access model, and variance contribution
3. A bias mitigation checklist -- practical rules for ensuring no
   single institution, region, or ideological framing dominates the
   corpus

---
QUALITY CRITERIA

Prefer sources that: publish on a predictable schedule, have stable
URL patterns suitable for automated collection, use named document
series (not just ad hoc publications), have open or
registration-only access (not hard paywalls), and have an archive of
at least 3 years. Flag but still include sources with mixed access
models. Do not exclude a source solely because it has an advocacy
stance -- bias diversity is itself a corpus value. Verify that each
source actually publishes downloadable files, not just HTML articles.
\end{promptbox}
\section{Question-Type and Persona Distribution}
\label{app:dist}
Beyond the question-type distribution (Table~\ref{tab:qtypes}), questions are
generated under 16 content-gated analyst personas, whose distribution reflects
which personas the document-profiling stage found supportable per report
(Table~\ref{tab:persona-dist}).

\begin{table}[h]
\centering\small
\begin{tabular}{lc}
\toprule
Persona & Share \\
\midrule
Geopolitical Researcher          & 19.1\% \\
Intelligence Analyst             & 11.0\% \\
Humanitarian Affairs Analyst     & 10.5\% \\
Risk Intelligence Analyst        & 10.4\% \\
Conflict Data Analyst            & 9.0\% \\
Financial Analyst                & 8.1\% \\
Military Analyst                 & 7.0\% \\
Economic Security Analyst        & 6.4\% \\
Strategic Communications Analyst & 4.1\% \\
Energy Market Analyst            & 3.1\% \\
Atrocity Crimes Analyst          & 2.5\% \\
Supply Chain Risk Analyst        & 2.3\% \\
OSINT Investigator               & 2.0\% \\
Cyber Threat Analyst             & 1.8\% \\
Disinformation Researcher        & 1.6\% \\
Sanctions Compliance Analyst     & 1.2\% \\
\bottomrule
\end{tabular}
\caption{Analyst-persona distribution over the external benchmark (16 personas).}
\label{tab:persona-dist}
\end{table}

\section{Example Q\&A Pairs}
\label{app:examples}
Table~\ref{tab:examples} gives one representative item per question type. Bracketed
markers (\texttt{[1]}) are citations into the retrieved-chunk set of the source
report; the cited chunks form the answer-cited gold used in
Section~\ref{sec:res-retrieval}.

\begin{table*}[t]
\centering\small
\begin{tabular}{p{2.0cm}p{5.2cm}p{6.4cm}}
\toprule
Type (persona) & Question & Reference answer \\
\midrule
\textbf{Factual} (Military Analyst) &
What specific types of decommissioned Soviet-era equipment were returned to service during 2023? &
Long-decommissioned equipment returned to service in 2023: tanks T-55 and T-62 [1]; howitzers D-10 and D-20 [1]. \\
\addlinespace
\textbf{Inferential} (Geopolitical Researcher) &
In the context of the India--China nuclear dyad, what is the implication of the ``stability--instability paradox''? &
The implication is that nuclear stability at the strategic level decreases the likelihood of nuclear confrontation, which in turn increases the chances of conventional confrontation; it is therefore suggested that India needs to hone its conventional capabilities [1]. \\
\addlinespace
\textbf{Comparative} (Intelligence Analyst) &
What technical advantage does the Jinan-1 microsatellite have over the Micius satellite? &
The Jinan-1 microsatellite weighs one-sixth of the Micius satellite and has 2--3 orders of magnitude better accuracy in quantum communication [1]. \\
\addlinespace
\textbf{Risk-oriented} (Geopolitical Researcher) &
What specific risk does the document associate with US bilateral deals with the DRC and Ukraine regarding critical minerals? &
The deals have been criticised as asymmetric, because potential security assistance is linked to access to resources [1]. \\
\bottomrule
\end{tabular}
\caption{Representative Q\&A pairs, one per question type.}
\label{tab:examples}
\end{table*}
\section{Judge Prompt}
\label{app:judge}
The LLM-as-judge scores each item on the six dimensions of
Section~\ref{sec:validation} against \emph{only} the retrieved chunks. We
reproduce the exact system prompt and the per-item user template below; the same
prompt is used for the human-validated selection and for the full-corpus scoring
(with \texttt{reasoning\_effort=low}).

\begin{promptbox}
[SYSTEM]
You are a strict senior intelligence analyst rating machine-generated
question-answer pairs for a retrieval benchmark, on a 1-3 scale, judging
ONLY against the retrieved chunks (never outside knowledge). For each
dimension: 3 = excellent/good, 2 = acceptable but flawed, 1 = poor or broken.
- groundedness: 3 the chunks clearly contain the answer; 2 partially/
  incomplete; 1 not answerable from chunks.
- doc_purpose_fit: 3 matches what the document is; 2 slightly stretched;
  1 category error (asks the doc for something its type cannot give).
- atomicity: 3 a single supported ask; 2 slightly overloaded; 1 fuses a
  supported part with an unsupported clause.
- clarity: 3 specific, self-contained, named referents; 2 vague or awkward;
  1 confusing/ambiguous.
- faithfulness: 3 answer fully supported by chunks with correct citations
  and no invented facts (a correct 'INSUFFICIENT_CONTEXT' counts as 3);
  2 minor unsupported detail; 1 invents facts/entities.
- overall: 3 a sound benchmark item; 2 usable but flawed; 1 discard/rewrite.
Return STRICT JSON only.

[USER]
Document: {doc_name}
Document type: {doc_type}
Analyst lens: {persona}
Question type: {question_type}

QUESTION:
{question}

RATIONALE:
{rationale}

ANSWER:
{answer}

RETRIEVED CHUNKS (only allowed evidence):
[1] {chunk_1}
[2] {chunk_2}
...

Return JSON: {"groundedness": int(1-3), "doc_purpose_fit": int(1-3),
"atomicity": int(1-3), "clarity": int(1-3), "faithfulness": int(1-3),
"overall": int(1-3), "note": "one short reason"}
\end{promptbox}
The RAG-reader evaluation (Table~\ref{tab:readers}) uses a separate, shorter judge
that scores a reader's answer on three dimensions against the gold context, with
the reference answer supplied only for the correctness comparison:

\begin{promptbox}
[SYSTEM]
You evaluate a model's ANSWER to an analyst question, judging ONLY against
the provided context passages (never outside knowledge), on a 1-3 scale.
- faithfulness: 3 every claim supported by the context, no invention;
  2 minor unsupported detail; 1 invents/contradicts.
- correctness: 3 correctly and completely answers the question from the
  context; 2 partially correct/incomplete; 1 wrong or non-answer.
  A correct 'INSUFFICIENT_CONTEXT' when the context truly lacks the answer
  = 3; an unjustified refusal = 1.
- overall: 3 excellent answer; 2 usable but flawed; 1 poor.
Return STRICT JSON only.

[USER]
Question: {question}

Context passages:
[1] {chunk_1}
[2] {chunk_2}
...

Reference answer (for correctness comparison):
{reference_answer}

MODEL ANSWER TO JUDGE:
{model_answer}

Return JSON: {"faithfulness": int(1-3), "correctness": int(1-3),
"overall": int(1-3), "note": "one short reason"}
\end{promptbox}
\section{Generation Prompts}
\label{app:genprompts}
The three generation stages of Section~\ref{sec:generation} (profiling,
question generation, answering) use the prompts below, served by Gemma.
Variables are shown as \{placeholders\}; excerpt lists are abbreviated.

\paragraph{Stage 1: Document profiling.}
\begin{promptbox}
[SYSTEM]
You profile intelligence/analytical documents to support downstream
question generation. Work STRICTLY from the supplied excerpts -- never
from outside knowledge. Your job is to characterise what kind of document
this is and, critically, what it can and cannot support being asked.
Hard rules:
- Name concrete actors, places, dates, organisations, programmes, figures,
  coding rules, methodologies and specific findings that actually appear.
- Do NOT include generic background, mission statements, 'X is a
  collaboration between...', boilerplate 'about the organisation' text,
  or download URLs.
- Distinguish what the document IS (e.g. a methodology codebook, a
  narrative/disinformation tracker, an analytical assessment, a dataset,
  a news report) from what it merely mentions.
- Be honest and explicit about limits: if the document defines how
  something is coded, that is descriptive/retrospective, not predictive;
  if it tracks how narratives spread, that is not independent verification
  of facts.
Return strict JSON.

[USER]
Document: {doc_name}
Excerpts gathered via multiple RAG probes ({n} unique chunks):
[1] section={section} | page={page}
{chunk_text}
...
Profile this document. Return JSON with EXACTLY these keys:
{
  "doc_type": short label for what this document is (e.g. "methodology
    codebook", "narrative/disinformation tracker", "analytical
    assessment", "statistical dataset", "news report"),
  "purpose": one sentence on what the document is actually for,
  "can_answer": [concrete question areas the excerpts genuinely support],
  "cannot_answer": [question areas that look topically adjacent but the
    document does NOT contain -- e.g. forecasts, market prices, sanctions
    lists, cross-episode comparisons, independent verification],
}
Do not invent. Do not add generic background.
\end{promptbox}

\paragraph{Stage 2: Question generation.}
\begin{promptbox}
[SYSTEM]
You generate benchmark questions for a retrieval/RAG evaluation over
intelligence documents. You see a document profile and the actual
representative excerpts. Generate only questions that the excerpts
genuinely support. Return strict JSON.
NON-NEGOTIABLE RULES:
1. ONE atomic ask per question. Never fuse a supported factual core with
   a second clause (a forecast, a comparison, or a domain application) --
   that is the single most common failure. If you catch yourself writing
   'X, and how might this affect Y', split it and keep only the supported
   half.
2. Gate the analyst lens on CONTENT, not identity. Only attach a lens
   (e.g. cyber, sanctions, finance/oil, military) if that domain genuinely
   appears in the excerpts. Do NOT force a persona's domain onto a document
   that never discusses it. A lens that has no anchor in the text must
   simply not be used.
3. Match the document's actual purpose. A methodology/codebook defines how
   things are coded -- never ask it for counts, country lists, or data
   values. A narrative/disinformation tracker records what is being said
   -- never ask it to verify facts, predict escalation, or quantify
   markets.
4. No forecasting/predictive framing ('next 30-90 days', 'early-warning
   indicators', 'anticipate escalation') unless the document is genuinely
   forward-looking. Descriptive or retrospective variables are NOT
   predictive indicators.
5. Comparative questions are welcome, but ONLY as within-document
   comparisons: both items being compared must each be explicitly present
   in the excerpts and individually answerable from them. NEVER compare to
   anything absent from the text -- no 'vs. previous conflicts / earlier
   versions / historical baselines / other datasets'. If only one side is
   present, ask a plain factual question instead.
6. No category substitution: repetition across outlets is NOT independent
   verification; narrative reach is NOT confirmed intent; a coding rule is
   NOT an early-warning signal.
7. Self-contained wording. Disambiguate every term inside the question
   itself (e.g. 'economic sectors', not just 'sectors'). Reference concrete
   entities/dates/figures from the excerpts.
8. Vary difficulty. Do not settle for surface facts only -- where the
   excerpts genuinely support it, include inferential, within-document
   comparative (per rule 5), and risk questions that require combining or
   reasoning across excerpts, while still obeying rules 1-6.
9. No near-duplicates, and this is strict. Two questions are duplicates if
   they would be answered by the same fact -- even when the wording differs,
   or when they differ only by one enumerated value. Before returning,
   re-read your own list and remove every question whose answer
   substantially overlaps another's. Each question must target a distinct
   fact.
10. Name the specific referent. When the document distinguishes several
    narratives, actors, or initiatives, name which one rather than asking
    about 'the narrative' in the abstract. Phrase every question in natural,
    plain language.
11. Rationale must match the question. The 'rationale' must explain why the
    exact information the question asks for matters -- not introduce broader
    context the question does not cover.

[USER]
Document: {doc_name} (collection={collection})
Document type: {doc_type}
Purpose: {purpose}
This document CAN support: {can_answer}
This document CANNOT support: {cannot_answer}

Representative excerpts (cite these indices as grounding):
[{idx}] section={section} | page={page}
{chunk_text}
...

Available analyst lenses (use ONLY those the excerpts actually support):
- {persona_name}: {persona_expertise}
...

Task: generate up to {max_questions} distinct benchmark questions that are
fully answerable from the excerpts above, following every non-negotiable
rule. Prefer fewer, well-grounded questions over hitting the cap. Span the
type mix where the content allows (factual, inferential, comparative,
risk_oriented) rather than defaulting to factual recall alone.
{optional type-specific hint for narrative or methodology documents}

Return JSON: {"questions": [{"question": str (atomic, self-contained),
"persona": one of the lens names above, "type": one of [factual,
inferential, comparative, risk_oriented], "rationale": why a senior analyst
asks this, "grounding": which excerpt indices support it}, ...]}
\end{promptbox}

\paragraph{Stage 3: Answering.}
\begin{promptbox}
[SYSTEM]
You answer analyst questions STRICTLY from the supplied excerpts.
- If the excerpts do not support an answer, reply exactly:
  'INSUFFICIENT_CONTEXT'.
- Cite chunk numbers in square brackets, e.g. [2], and make sure each
  citation points to the excerpt that actually supports that specific
  claim -- do not cite a chunk that does not contain the stated fact.
- Do not import outside knowledge or assume facts the excerpts omit.
- Name only entities, actors, and figures that appear explicitly in the
  excerpts; never add a name from general knowledge that is not in the
  cited chunks.
(For inferential/risk questions, one rule is appended:)
- This is an inferential/risk question: mark every inference with modal
  qualifiers ('may', 'could', 'might') and never upgrade a mentioned actor
  into holding a position the excerpts do not attribute to them.

[USER]
Analyst lens (for tone & emphasis only):
{persona}

Question:
{question}

Excerpts:
[1] {chunk_text}
[2] {chunk_text}
...
\end{promptbox}
\section{Embedding Query/Passage Prefixes}
\label{app:embprefix}
Each embedding family is applied with its recommended instruction prefix;
BM25 uses none. We follow the published formats:

\begin{table}[h]
\centering\footnotesize
\setlength{\tabcolsep}{4pt}
\renewcommand{\arraystretch}{1.15}
\begin{tabular}{@{}lp{3.2cm}l@{}}
\toprule
Family & Query prefix & Passage \\
\midrule
E5 / mE5        & \texttt{query:}    & \texttt{passage:} \\
BGE (en)        & \texttt{Represent this sentence for searching relevant passages:} & --- \\
BGE-M3          & ---                 & --- \\
Qwen3-Emb       & \texttt{Instruct: \ldots{} Query:} & --- \\
GTE, MiniLM     & ---                 & --- \\
OpenAI, Mistral & ---                 & --- \\
\bottomrule
\end{tabular}
\caption{Per-family embedding prefixes. The Qwen3 query instruction in full is
``Instruct: Given a web search query, retrieve relevant passages that answer
the query\textbackslash nQuery:\,''.}
\label{tab:embprefix}
\end{table}
\end{document}